\documentclass[runningheads]{llncs}
\usepackage[T1]{fontenc}
\usepackage{graphicx}

\usepackage{doi}
\usepackage{multirow}
\usepackage{tabularx}
\usepackage{booktabs}
\usepackage{xcolor}
\usepackage{comment}
\usepackage{color}
\usepackage{graphicx}
\usepackage{textcomp}
\usepackage{xcolor}
\usepackage{dsfont}
\usepackage{siunitx}
\newcommand{\NA}{\multicolumn{1}{c}{--}}
\begin{document}
\title{SleepNet and DreamNet: Enriching and Reconstructing Representations for Consolidated Visual Classification}
%
%
\author{Mingze Ni\inst{1}\orcidID{0000-0003-1220-365X} \and
Wei Liu\inst{1}\orcidID{0000-0002-3003-1313}}
\authorrunning{M. Ni and W.Liu}
%
\institute{University of Technology Sydney, 15 Broadway, Ultimo, 2007 Australia \email{mingze.ni@uts.edu.au} ; \email{wei.liu@uts.edu.au}
}
\maketitle              
\begin{abstract}
An effective integration of rich feature representations with robust classification mechanisms remains a key challenge in visual understanding tasks. This study introduces two novel deep learning models, SleepNet and DreamNet, which are designed to improve representation utilization through feature enrichment and reconstruction strategies. SleepNet integrates supervised learning with representations obtained from pre-trained encoders, leading to stronger and more robust feature learning. Building on this foundation, DreamNet incorporates pre-trained encoder–decoder frameworks to reconstruct hidden states, allowing deeper consolidation and refinement of visual representations. Our experiments show that our models consistently achieve superior performance compared with existing state-of-the-art methods, demonstrating the effectiveness of the proposed enrichment and reconstruction approaches.

\end{abstract}
\section{Introduction}
In the current digital age, the ability to accurately classify large visual datasets has become of paramount importance across a myriad of fields such as computer vision (CV) \cite{alexnet} \cite{VGG} \cite{googlenet} \cite{vit}. The blossoming of artificial intelligence and deep learning has greatly facilitated the handling of complex classification tasks. Deep learning's capacity to sift through multitudes of variables, discern patterns, and extract key features has led to impressive breakthroughs in numerous applications, from image recognition to disease prediction \cite{notbugs}.

The groundbreaking convolutional neural networks (ConvNets), such as ResNet \cite{resnet} and EfficientNet \cite{tan2019efficientnet}, have emerged as dominant architectures in computer vision. ResNet addresses the vanishing gradient issue through deep residual networks, enabling deeper models without performance loss, while EfficientNet introduced a compound scaling method that scales depth, width, and resolution, enhancing both efficiency and accuracy. These models have set new benchmarks across various datasets and have been pivotal in applications such as autonomous driving and advanced image recognition, reshaping how machines interpret visual data. Meanwhile, the success of pre-trained Transformer architectures \cite{vaswani2017attention} like ViT \cite{vit} for vision tasks has shown that using primarily standard Transformer layers can achieve significant performance in downstream applications, reaching levels comparable to previous state-of-the-art neural networks and suggesting that Transformers may offer greater scalability across diverse domains.

Transformers have demonstrated superior model capabilities but often suffer from poor generalization when compared to chain-like networks due to a lack of appropriate inductive bias \cite{wu2021cvt}. Recent research has focused on hybrid methods that combine the structures of both to retain their respective advantages \cite{resnet-vit} \cite{wu2021cvt} \cite{dai2021coatnet}. For example, Convolution Vision Transformers (CvT) \cite{wu2021cvt} enhance performance by integrating convolutional token embedding and convolutional transformer blocks with convolutional projection, aiming for improved accuracy and efficiency. Similarly, the convolution and attention transformer (CoAtNet) \cite{dai2021coatnet} boosts performance by introducing relative attention that merges convolution and attention mechanisms, enhancing generalizability and efficiency through a simple stacking of convolution and attention layers.

However, these models still lack generalizability as their evaluations are often limited to specific architectures and focus primarily on testing the architecture superficially, without delving into deeper conceptual explorations, potentially leading to a lack of broad applicability. In response to this challenge, we propose two innovative networks designed for visual classification tasks. The first, SleepNet, introduces a novel learning paradigm that incorporates \textit{sleep blocks} (which we shall explain later) into the training process of neural networks. The name ``SleepNet" is inspired by the concept of memory consolidation during sleep, where information is processed and integrated. This model integrates features from pre-trained encoders into designated layers, creating periods of feature enrichment interspersed within supervised learning epochs. Notably, SleepNet preserves the weights of the pre-trained encoding components in each enrichment connection during the supervised phases, promoting stable representation learning. Building on this foundation, we developed DreamNet, which enhances SleepNet by introducing a \textit{reconstruction block}. The name ``DreamNet" draws a parallel to the idea of ``dreaming" as a process of refining and generating new information. DreamNet utilizes a complete pre-trained autoencoder, not just the encoder, to deepen feature consolidation and to generate new information. This autoencoder not only reconstructs the hidden states but also serves as an additional feature enhancer compared to SleepNet, thereby boosting overall performance.

This research contributes significantly to the field of deep learning in the following crucial ways:
\begin{itemize}
    \item We present an innovative deep learning strategy, \textbf{SleepNet}, which enhances deep learning processes through a novel feature enrichment mechanism. SleepNet combines pre-trained encoders with supervised learning in a hybrid framework, enabling rich data interpretation and robust supervised predictions.
    \item Building on SleepNet, we introduce \textbf{DreamNet}, which leverages a pre-trained encoder-decoder framework for advanced feature augmentation and information reconstruction. This approach enhances knowledge transfer and consolidates training, resulting in further performance improvements.
    \item Both SleepNet and DreamNet have significant potential for general applicability within computer vision. We introduce versions of these models specifically tailored for image classification to demonstrate their capabilities. Our extensive experiments demonstrate that the proposed methods, especially DreamNet, consistently outperform state-of-the-art baselines on diverse image classification tasks, highlighting their superior efficacy and potential for universal applicability.
\end{itemize}

\section{Related Work}\label{related work}
Deep learning has been significantly advanced with the development of supervised learning models like ResNet \cite{resnet} and MobileNet \cite{MobileNets}, which excel in multiple tasks across image processing and object detection. These models share a common architecture: chain-like structures where components are sequentially linked, facilitating a systematic enhancement of feature extraction. For instance, ResNet utilizes residual blocks connected by skip connections to tackle the vanishing gradient problem in deep networks, while MobileNet is optimized for mobile environments with streamlined CNN structures. Although highly effective in recognizing patterns from vast labeled datasets, this reliance on extensive labeled data can limit their adaptability and generalization in new, less structured environments.

Pre-trained models, such as Variational Autoencoders (VAEs) \cite{vae} and Generative Adversarial Networks (GANs) \cite{gan}, play a key role in understanding the inherent structure and distribution of data. These models are often pre-trained on large datasets to learn rich representations, with applications ranging from image generation to style transfer. While effective for representation learning, their direct application to task-specific classification can be limited. In contrast, Transformer architectures \cite{vaswani2017attention}, particularly with their self-attention capabilities, have revolutionized various domains. Vision Transformer (ViT) \cite{vit} extends this approach to image classification, achieving remarkable results after pre-training on large-scale datasets, although it struggles with limited data and generality \cite{wu2021cvt}.

The integration of convolutional layers with pre-trained models, especially transformers, has led to several innovative hybrid architectures, blending the strengths of CNNs and transformers to enhance feature extraction and understanding of local relationships. Notable implementations include replacing multi-head attention with convolution layers or adding them sequentially or in parallel to transformer blocks. Specific examples like Convolution Vision Transformers (CvT) \cite{wu2021cvt} utilize convolutional projections instead of traditional position-wise linear projections for attention mechanisms, while convolution and attention transformer (CoAtNet) \cite{dai2021coatnet} combines convolution and attention layers to improve performance and efficiency. Additionally, hierarchical multi-stage structures similar to CNNs have been introduced, significantly boosting performance. These methods augment traditional ConvNets with self-attention modules or incorporate convolutional properties directly into transformer backbones such as ResNet-ViT \cite{resnet-vit}, demonstrating the potential of these hybrid models to leverage the strengths of both paradigms effectively.

Deep learning seeks to improve efficiency and generalization by consolidating learned representations. Researchers are actively exploring iterative feature refinement \cite{Jaderberg2017}, knowledge transfer \cite{Hinton2015}, and hierarchical representation learning \cite{Bengio2013}. We explore computational strategies for enhancing deep learning models by framing information transfer and reconstruction as opportunities for feature augmentation and representation regularization. The following sections introduce methods that leverage these principles to improve task performance, reduce computational complexity, and enhance model robustness.

\section{Methodology}\label{methodology}

In general, deep neural networks are constructed by connecting many weight matrices and nonlinear operators. In this paper, we consider a chain-like neural network constructed by stacking similar deep neural blocks, such as Multilayer Perceptron or ResNet. Let $D={(x_1,y_1),(x_2,y_2),\ldots,(x_n,y_n)}$ be a dataset consisting of $n$ samples, where $x$ and $y$ represent the input image and its corresponding class, respectively. Given a chain classifier with $M$ layers, $C(x)=\left(g_M \circ g_{M-1} \circ \ldots \circ g_2 \circ g_1\right)(x)$ maps from the image space $\mathcal{X}$ to the $k$ classes, where $g_{m}(\cdot)$ is the $m$th neural block. The output of the $m$th block is $h_m(x)=g_m(h_{m-1}(x))$, and the neural blocks can be similar blocks constructed by fully connected layers, convolutional layers, pooling and normalizing layers with the activation functions. To construct a SleepNet, we also need a pre-trained autoencoder component $P=\theta(\phi(x))$ where $\theta(\cdot)$ and $\phi(\cdot)$ are the encoder and decoder, respectively. The hidden state from the latent space is $a=\phi(x)$.

\subsection{SleepNet}
The proposed method, SleepNet, incorporates a pre-trained encoder component to feed input and hidden states, producing enriched encodings for subsequent layers. This fusion of information inside a single architecture leverages the concept of iterative feature enrichment and consolidation, and applies it to machine learning tasks.

\subsubsection{Enrichment Connection}
SleepNet is designed to harness the strengths of pre-trained encoders into supervised learning models, consolidating them into one cohesive method. Its unique attribute incorporates a pre-trained encoder $\phi(\cdot)$, derived from an autoencoder framework. The reason for only utilizing the encoder is that a well-trained encoder is expected to transform more features to the input \cite{pretrained}, and excluding the decoder will reduce the complexity of the proposed method. We contend that such an encoder setup amplifies feature extraction, easing the learning process.

SleepNet uses a pre-trained encoder \( \phi \) to process hidden states \( h_i \), forwarding encoded features to subsequent blocks through a mechanism that we call ``enrichment connection.'' This approach harnesses latent features for the supervised learning process. ``Sleep blocks'' interspersed within the model bridge the output of supervised blocks \( h_m = g_m(h_{m-1}) \) with encoded features \( \phi(h_{m-1}) \), enhancing the integration of pre-trained insights into the learning sequence. The ``sleep block'' \( s(\cdot) \) is mathematically expressed as:
\begin{equation}
    s_m(\mathbf{h}_m)=g_m(\mathbf{h}_{m-1})+\phi(\mathbf{h}_{m-1}) \label{eqt: connection},
\end{equation}
with \( \mathbf{h}_{m-1} \) denoting the output from the \( m \)th sleep block. The supervised blocks capture local details, whereas encoder \( \phi \) elevates local to non-local information, facilitating enhanced feature fusion.

\begin{figure*}[t!]
    \centering
    \includegraphics[width=0.8\textwidth]{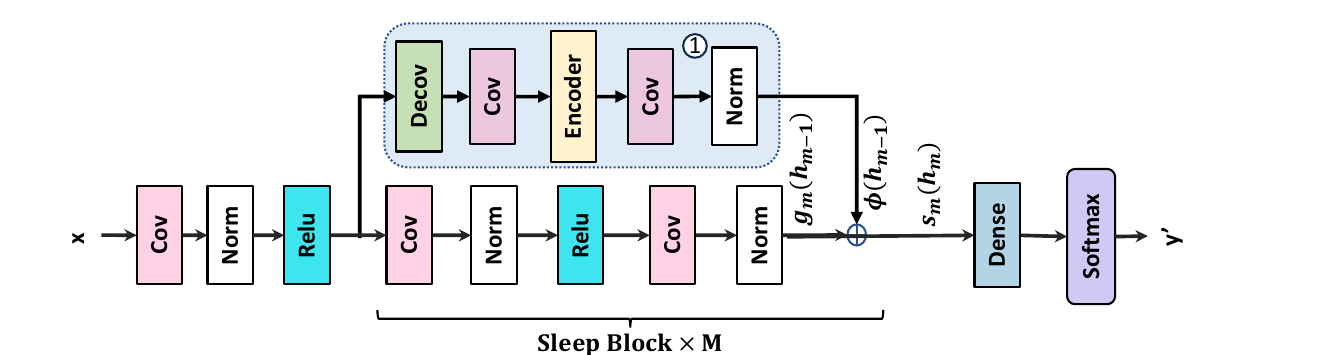}
    \caption{Overview of the Visual SleepNet Architecture, featuring M ``sleep blocks'' that are constructed by chain-like blocks processing data through convolutional layers (``Cov''), normalization (``Norm''), and an enrichment connection in block \textcircled{1}. The enrichment connection includes an encoder and a deconvolution layer for feature extraction and dimension adjustment, facilitating robust representation formation. The workflow culminates in a dense layer followed by a softmax classifier for output classification.}
    \label{fig: cv sleepnet}
\end{figure*}

In computer vision applications, SleepNet begins by processing an input image through convolutional (``Cov'') layers to extract preliminary features, which are then normalized (``Norm'') and activated through ReLU functions to prepare them for further processing. The core of SleepNet consists of sleep block, each consisting of an enrichment connection and chain-like blocks. The enrichment connection (Block \textcircled{1} in Fig \ref{fig: cv sleepnet}) adjusts the feature dimensions via deconvolutional (``Decov'') and convolutional layers to match the fixed input size required by the typically pre-trained encoder, \(\phi(\cdot)\). This is essential because initial processing often reduces input dimensionality, potentially mismatching the encoder’s specifications. After dimension adjustment and feature enhancement by the encoder, further convolutional and normalization layers refine and compress the features. Simultaneously, the initial output is processed through chain-like blocks for independent feature extraction. Outputs from both pathways are then merged by simple addition, repeating across multiple sleep block to progressively enhance features. This integrated workflow in Fig \ref{fig: cv sleepnet} culminates in a dense layer, leading to a softmax classifier for final image classification, effectively combining features from pre-trained components with supervised learning to improve prediction accuracy.

\subsection{DreamNet}
Building on the SleepNet framework, DreamNet is designed to simulate and leverage processes of representation reconstruction. It enhances existing enrichment connections with a novel ``reconstruction connection'', utilizing a full pre-trained autoencoder setup. Additionally, DreamNet learns augmented features by reconstructing hidden states using an autoencoder, akin to how a system might refine its internal models based on recalled information.

\subsubsection{Reconstruction Connection}
We propose a ``reconstruction connection'' to enhance the previously proposed ``enrichment connection'' through two innovative modifications. Firstly, we replace the encoder \(\phi(.)\) with a complete pre-trained autoencoder \(\theta(\phi(x))\). We anticipate that utilizing the full autoencoder will allow for a more comprehensive integration of information into the subsequent block. The mathematical illustration of a ``reconstruction connection'' is as follows:
\begin{equation}
    s_m(\mathbf{h}_m)=g_m(\mathbf{h}_{m-1})+\theta(\phi(\mathbf{h_{m-1}})) \label{eqt: dream connection},
\end{equation} where $\theta(\cdot)$ and $\phi(\cdot)$ are encoder and decoder, respectively. Secondly, the output (``reconstructed representation'') generated by the autoencoder \(\theta(\phi(x))\) is not only transferred to the subsequent block but also directed to a parallel block for a deeper analysis of these reconstructed features. In the final stage, the data from the reconstructed representations and the chain-like module is combined before being processed through fully connected layers, ultimately leading to a softmax layer for predictions.

DreamNet is expected to achieve better performance by utilizing the pre-trained autoencoder for feature augmentation, which is theoretically and practically different from traditional data augmentation. Specifically, conventional augmentation methods, such as RandAugment \cite{randaugment}, typically involve crude manipulations on the raw dataset. While these methods can generate similar data, they may also introduce unnecessary noise. In contrast, feature augmentation through ``reconstruction'' is a more principled approach that leverages the model's architecture. Practically, data augmentation and feature augmentation can be used complementarily within the same model, where data augmentation operates on the raw data, while our proposed feature augmentations with ``reconstruction'' are based on enhancing the model's internal representations.

\begin{figure*}[t!]
    \centering
    \includegraphics[width=0.8\textwidth]{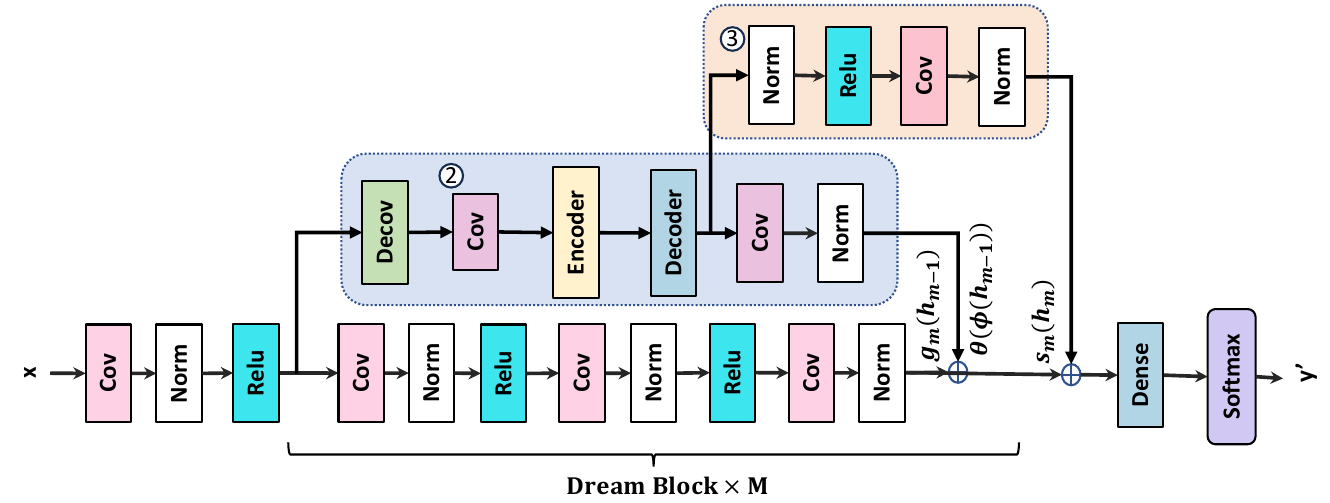}
    \caption{Overview of the DreamNet architecture, integrating ``Dream Blocks'' that include chain-like blocks processing data through convolutional layers (``Cov''), normalization (``Norm''), and the reconstruction connection, enhanced with a full encoder-decoder setup for advanced feature consolidation and reconstruction. The reconstructed representations will be processed in block \textcircled{3} with convolutional layers and pass to the dense layer for the final classification.}
    \label{fig: cv dreamnet}
\end{figure*}

\begin{figure}
    \centering
    \includegraphics[width=0.95\columnwidth]{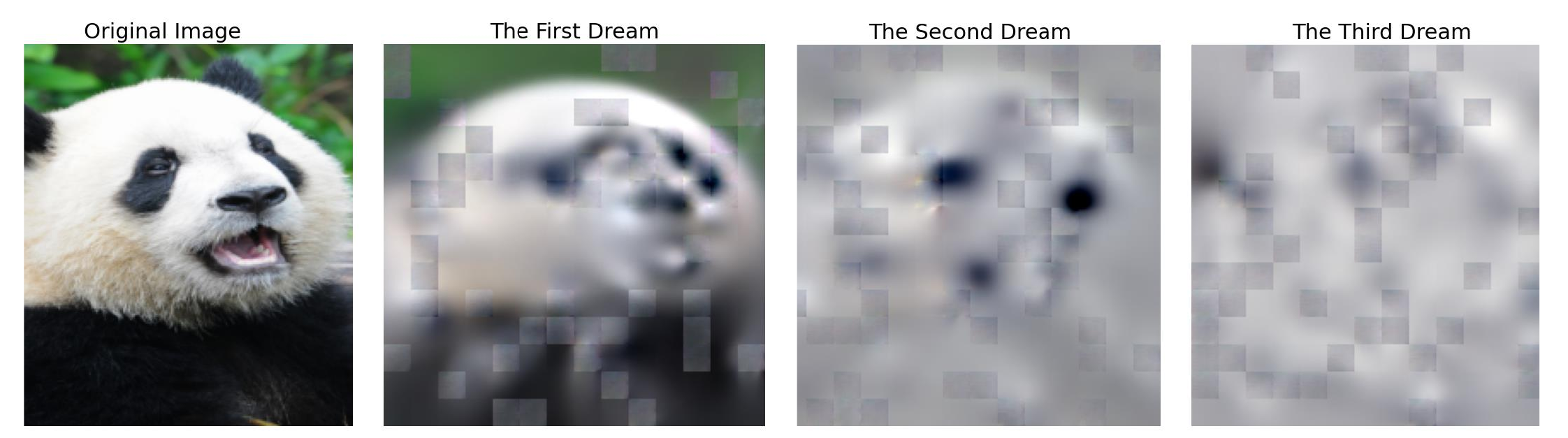}
    \caption{Stages of image transformation by DreamNet-3 using a masked autoencoder (MAE) \cite{mae}. Starting with the `Original Image' of a panda \cite{panda}, the sequence through ``The First Reconstruction'', ``The Second Reconstruction'', and ``The Third Reconstruction'' illustrates progressive abstractions, depicting the model's process of deepening feature exploration and refinement.}
    \label{fig: cv dreams}
\end{figure}

DreamNet enhances SleepNet’s architecture for computer vision by incorporating a ``reconstruction connection'' that employs a comprehensive autoencoder for advanced feature consolidation, as illustrated in Fig \ref{fig: cv dreamnet}. DreamNet introduces ``Dream Blocks'' which not only include standard chain-like blocks for initial feature extraction but also feature an advanced reconstruction connection setup (block \textcircled{2} in Fig \ref{fig: cv dreamnet}). This reconstruction connection uses a full autoencoder that first encodes the feature data to a latent, compressed representation and then decodes it, effectively reconstructing and enhancing the image data. This reconstructed output undergoes further refinement in block \textcircled{3} in Fig \ref{fig: cv dreamnet}, where additional convolutional and normalization layers refine these features to ensure robust feature extraction. This continuous cycle of encoding, decoding, and refinement significantly bolsters the model's pattern recognition capabilities. The workflow culminates in a dense layer followed by a softmax classifier, ensuring precise classification by effectively leveraging the enhanced feature set. To give a more valid illustration, we present Fig \ref{fig: cv dreams} by plotting the original image and the generated reconstructed images.

\section{Experiments}\label{experiments}
In this section, we provide comprehensive empirical evaluations on our proposed SleepNet and DreamNet models using real-world datasets against other state-of-the-art models.
To evaluate SleepNet and DreamNet, we conducted experiments on CIFAR100 \cite{cifar100}, ImageNet-tiny \cite{imagenettiny}, and ImageNet 1K \cite{imagenet}, varying the number of enrichment/reconstruction blocks (\(M\)) from 1 to 4, denoted as SleepNet-1 to SleepNet-4 and DreamNet-1 to DreamNet-4. For fair comparison, all models were trained under the same settings for 30 epochs using ADAM \cite{adam} with \(lr = 0.005\), \(\beta_1 = 0.9\), \(\beta_2 = 0.999\), and \(\sigma = 10^{-5}\). The implementation is publicly available on GitHub\footnote{\url{https://github.com/MingzeLucasNi/DreamNet-SleepNet.git}}.


\subsection{Augmentation Methods}
Data augmentation techniques are crucial for enhancing model performance in computer vision tasks and are widely applied to state-of-the-art visual models. To ensure robust comparisons, we incorporated two powerful data augmentation methods, namely RandAugment \cite{randaugment} and Mixup \cite{zhang2017mixup}, into our visual models and the baselines. RandAugment systematically searches for the best augmentation policies by randomly applying a fixed number of distortions, optimizing the augmentation strategy for better generalization. Mixup, on the other hand, creates new training examples by combining pairs of images and their labels, encouraging models to behave linearly in between training examples and improving robustness against adversarial examples.

\subsection{Datasets and Metrics}
\subsubsection*{Datasets}
To ensure thorough evaluation, we used diverse datasets for visual tasks, including CIFAR-100 \cite{cifar100}, ImageNet-tiny \cite{imagenettiny}, and ImageNet-1K \cite{imagenet}. This selection ensures a comprehensive assessment, highlighting the versatility and effectiveness of SleepNet and DreamNet in visual tasks. Dataset specifics are in Table \ref{tab: datasets}.

\subsubsection*{Metrics}
To evaluate model performance, we use classification accuracy, model size, and complexity, measured by input size, number of parameters, and FLOPs. \textbf{Accuracy} is defined as the ratio of correct predictions to total predictions, \textbf{input size} refers to the dimensions of images fed into the model, \textbf{\#Params} denotes the total number of trainable weights, and \textbf{FLOPs} represent the number of floating-point operations required per prediction.


\subsection{Baselines}
During the evaluation of our models, we benchmarked their performance against various state-of-the-art models to ensure comprehensive comparisons. For vision-focused tasks, we utilized:
\begin{itemize}
    \item \textbf{ResNet variants} \cite{resnet}: ResNet18 and ResNet50, known for their deep residual learning capabilities in image classification.
    \item \textbf{EfficientNet} \cite{tan2019efficientnet}: Models that balance network depth, width, and resolution for optimal performance using compound scaling.
    \item \textbf{Attention-based Transformers}: Vision Transformer (ViT) \cite{vit}, leveraging self-attention for processing image patches.
    \item \textbf{Hybrid Methods}: CoAtNet \cite{dai2021coatnet} and CvT \cite{wu2021cvt}, combining CNNs with attention mechanisms for efficient image processing.
    \item \textbf{Augmentation-based Models}: Augmenting Convolutional Network (ACN) \cite{2021augmentingACN} and Masked Augmentation Model (MAS) \cite{heo2024maskingmcs}, enhancing feature selection during training.
\end{itemize}

\begin{table}[t]
\small
\centering
\caption{Overview of datasets used in the experiments}
\begin{tabular}{rrrr}
\toprule
Dataset ~&~Training Size ~&~Testing Size~      & ~ \#Classes ~        \\
\midrule
CIFAR100&  50,000   &  10,000   &  100  \\
ImageNet-tiny& 100,000     &  1,000   &  200  \\
ImageNet-1k& 1,281,167     &  100,000   &  1,000  \\

\bottomrule
\end{tabular}  

\label{tab: datasets}
\end{table}

\subsection{Main Results and Analysis}
The main results of the experiments are presented in Tables  \ref{tab: cv model_performance}. 
In the evaluation of computer vision tasks, vision transformers, particularly \(\text{MAE-large}\), demonstrated strong performance, achieving \(91.3\%\) on CIFAR100 and \(87.1\%\) on ImageNet-tiny. Hybrid models, such as \(\text{CvT-W24}\) and \(\text{CoAtNet-3}\), also performed impressively across all datasets, highlighting the benefits of integrating convolutional and attention mechanisms. Notably, \(\text{DreamNet-3}_{\text{MAE-l}}\) achieved the highest accuracy on CIFAR100 (\(93.4\%\)) and Tiny-ImageNet (\(89.6\%\)), while \(\text{DreamNet-4}\) recorded the best results on ImageNet 1K (\(89.2\%\)). Additionally, SleepNet consistently outperformed the baselines, securing the second-best results and displaying a performance closely competitive with that of DreamNet.

The superior performance of DreamNet over SleepNet and other models can be attributed to several factors. The dream connection in DreamNet refines and consolidates features learned during training, enhancing the model's ability to generalize to new data. This mechanism allows DreamNet-3 to achieve the highest accuracy on ImageNet 1K (87.8\%) and CIFAR100 (92.3\%). Additionally, the use of pre-trained self-supervised encoders in both SleepNet and DreamNet provides a strong initialization, further improving their feature extraction capabilities. \(\text{DreamNet-3}_{\text{MAE-l}}\) set a new benchmark on CIFAR100 with an accuracy of 93.4\%, highlighting the significant improvements brought by the reconstruction connections in enhancing learning and generalization.

\begin{table}[t]
\centering
\caption{Performance comparison of SleepNet, DreamNet, and baseline models on vision tasks. Metrics include input size, parameter count, FLOPs, and accuracy. Best results are highlighted in bold.}
\label{tab: cv model_performance}
\resizebox{\linewidth}{!}{%
\begin{tabular}{@{}p{2cm}p{2.5cm}p{1.8cm}S[table-format=3.2]S[table-format=5.1]S[table-format=2.1]S[table-format=2.1]S[table-format=2.1]@{}}
\toprule
\textbf{Type} & \textbf{Model} & \textbf{\shortstack{Input Size}} & {\#Params} & {FLOPs} & {\shortstack{CIFAR100}} & {\shortstack{ImageNetTiny}} & {\shortstack{ImageNet1K}} \\ 
\midrule

\multirow{5}{*}{\shortstack{Conv only}} 
& EfficientNet-B7 & $600^2$ & 66.44 & 37.0 & 90.1 & 80.1 & 84.7 \\
& EfficientNetV2-L & $480^2$ & 121.32 & 53.0 & 92.1 & 77.3 & 85.7 \\
& ResNet18 & $224^2$ & 11.23 & 1.8 & 80.6 & 68.9 & 69.7 \\
& ResNet50 & $224^2$ & 25.12 & 3.8 & 86.9 & 68.0 & 76.0 \\
& ResNet101 & $224^2$ & 45.01 & 7.6 & 84.1 & \NA & 80.8 \\

\midrule
\multirow{4}{*}{Transformer} 
& \(\text{ViT}_{\text{base}}\) & $224^2$ & 86.45 & 55.4 & 87.3 & 86.1 & 79.4 \\
& \(\text{ViT}_{\text{large}}\) & $384^2$ & 307.12 & 190.7 & 91.0 & 88.0 & 83.0 \\
& \(\text{MAE}_{\text{base}}\) & $224^2$ & 86.43 & 17.6 & 91.1 & 87.0 & 82.6 \\
& \(\text{MAE}_{\text{large}}\) & $224^2$ & 304.18 & 61.9 & 91.3 & 87.1 & 83.1 \\

\midrule
\multirow{4}{*}{\shortstack{Conv + Trans}} 
& CvT-21 & $384^2$ & 32.41 & 24.9 & 90.1 & 83.1 & 83.3 \\
& CvT-W24 & $384^2$ & 277.33 & 193.2 & 92.1 & 88.1 & \NA \\
& CoAtNet-2 & $224^2$ & 75.14 & 15.7 & \NA & 87.1 & 84.1 \\
& CoAtNet-3 & $224^2$ & 168.12 & 34.7 & \NA & 87.6 & 84.5 \\

\midrule
\multirow{2}{*}{\shortstack{Augmentation\\+Trans}} 
& \(\text{ViT}_{\text{l}}\)-ACN & $384^2$ & 490.11 & 41.9 & 91.2 & \NA & 85.7 \\
& \(\text{MAE}_{\text{l}}\)-MAS & $224^2$ & 551.42 & 29.9 & 90.0 & \NA & 83.6 \\

\midrule
\multirow{8}{*}{SleepNet} 
& SleepNet-2 & $224^2$ & 10.01 & 40.5 & 83.1 & 64.2 & 63.0 \\
& SleepNet-3 & $224^2$ & 10.07 & 60.6 & 92.2 & 88.1 & 85.9 \\
& SleepNet-4 & $224^2$ & 10.14 & 80.7 & 91.1 & 88.2 & 83.9 \\
& \(\text{SleepNet-3}_{\text{ViT-b}}\) & $224^2$ & 10.07 & 107.3 & 92.2 & 88.1 & 85.9 \\
& \(\text{SleepNet-3}_{\text{ViT-l}}\) & $224^2$ & 13.40 & 371.5 & 92.3 & 88.4 & 86.4 \\
& \(\text{SleepNet-3}_{\text{MAE-b}}\) & $224^2$ & 10.07 & 60.6 & 90.2 & 86.1 & 81.9 \\
& \(\text{SleepNet-3}_{\text{MAE-l}}\) & $224^2$ & 10.09 & 126.5 & 91.0 & 86.3 & 84.1 \\

\midrule
\multirow{5}{*}{DreamNet} 
& DreamNet-2 & $224^2$ & 208.31 & 52.3 & 85.1 & 71.9 & 71.6 \\
& DreamNet-3 & $224^2$ & 209.12 & 83.5 & 92.3 & 89.1 & 87.8 \\
& DreamNet-4 & $224^2$ & 212.03 & 118.4 & 85.1 & 89.9 & {\bfseries 89.2} \\
& \(\text{DreamNet-3}_{\text{MAE-b}}\) & $224^2$ & 209.12 & 83.5 & 92.3 & 89.1 & 87.8 \\
& \(\text{DreamNet-3}_{\text{MAE-l}}\) & $224^2$ & 209.90 & 149.4 & {\bfseries 93.4} & {\bfseries 89.6} & 88.9 \\

\bottomrule
\end{tabular}%
}
\end{table}

\subsection{Ablation Studies} \label{ablation}

\subsubsection{Performance comparison between DreamNet and SleepNet.}

Table \ref{tab: cv model_performance} shows that DreamNet consistently outperforms SleepNet across different datasets, demonstrating its stronger effectiveness for visual classification tasks. This superior performance can be explained by several architectural advantages. First, the reconstruction blocks in DreamNet iteratively refine and consolidate learned representations, helping the model capture more detailed and discriminative patterns while improving generalization. Second, DreamNet leverages pre-trained encoders, which provide a stronger initialization and enable more robust feature extraction from complex visual data. These pre-trained representations give DreamNet a more informative starting point than training from scratch, allowing it to learn more effectively. In addition, the interaction between the reconstruction blocks and the encoder features further enhances representation learning, leading to better classification accuracy. Together, these factors account for DreamNet’s consistent advantage over SleepNet and other baseline models.

\subsubsection{Comparisons with different numbers of enrichment and reconstruction blocks}
We also examine how the number of enrichment and reconstruction blocks affects the performance of SleepNet and DreamNet. As shown in Table \ref{tab: cv model_performance}, increasing the number of blocks generally leads to better accuracy, indicating that deeper variants are more effective at learning discriminative representations. On CIFAR100, both SleepNet-4 and DreamNet-4 achieved clear improvements over shallower versions, with DreamNet-4 reaching 92.3\% accuracy, while on ImageNet-tiny DreamNet-4 again delivered the best performance among all configurations. This trend suggests that adding more blocks enables the models to capture richer hierarchical features and more complex visual patterns, which is especially important for challenging image classification tasks. Moreover, because both architectures build on pre-trained encoders, the added blocks do not simply increase depth, but also progressively refine already strong feature representations. DreamNet’s consistent advantage over SleepNet further shows that the reconstruction mechanism is particularly effective in consolidating and enhancing learned features, making deeper DreamNet variants especially powerful across datasets.

\subsubsection{Comparisons with different chain-like models.}
\begin{figure}[t!]
    \centering
    \includegraphics[width=0.9\columnwidth]{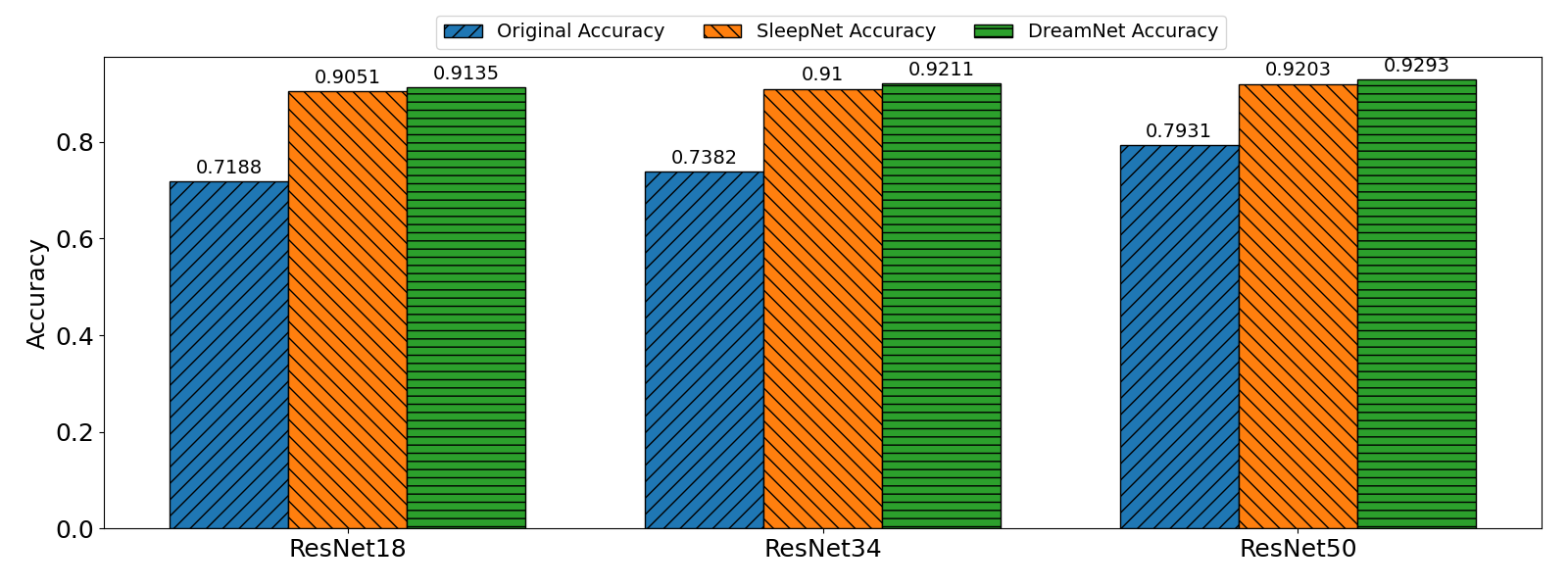}
    \caption{Ablation studies for testing the different chain-like blocks by comparing the original performance  of various vision classifiers against their performance when integrated with the proposed methods, SleepNet and DreamNet.}
    \label{fig: ablation sup}
\end{figure}

We evaluated the efficacy of various supervised models using distinct chain-like blocks: ResNet18, ResNet34, and ResNet50, which vary in the number of convolutional layers. Fig \ref{fig: ablation sup} summarizes their performance on the CIFAR100 dataset. There is a clear trend that shows that the better the performance of the chain-like block, the better the subsequent performance of the proposed models (SleepNet and DreamNet). This correlation suggests that stronger initial performance from a classifier enhances its subsequent integration into SleepNet, likely due to the foundational role of chain-like blocks in the models.

\begin{figure}
    \centering
    \includegraphics[width=0.9\columnwidth]{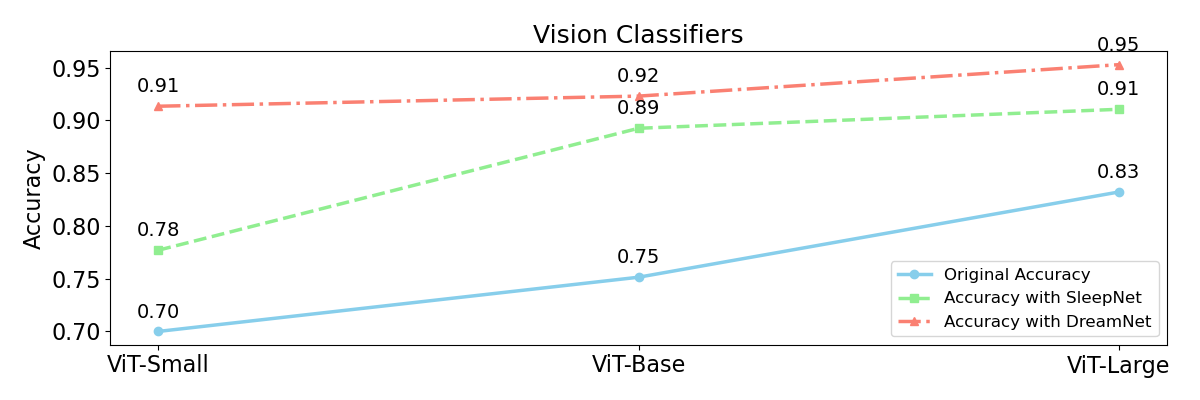}
    \caption{Ablation studies for testing the different pre-trained encoders/autoencoders by comparing the original performance of various vision classifiers against their performance when integrated with the proposed methods, SleepNet and DreamNet.}
    \label{fig: ablation unsup}
\end{figure}

\subsubsection{Comparisons with different unsupervised encoders and autoencoders.}

We assessed the performance of SleepNet and DreamNet by utilizing various pre-trained encoders and autoencoders, as depicted in Fig \ref{fig: ablation unsup}. Incorporating more sophisticated encoders consistently improved results: employing ViT-Small with SleepNet increased accuracy from 0.70 to 0.78, and with DreamNet, it reached 0.91. This trend demonstrates that the more advanced the encoder, the better the performance of the proposed methods. The primary reason for this improvement is that sophisticated encoders provide richer feature representations, enhancing the learning process. SleepNet benefits from the additional feature extraction capabilities, while DreamNet's reconstruction blocks further refine and consolidate these features, leading to superior performance. This highlights the importance of utilizing robust, pre-trained encoders to augment the overall efficacy of these models.

\subsubsection{Comparisons with frozen and unfrozen unsupervised encoders and autoencoders.}

\begin{figure}[t!]
\centering
\includegraphics[width=0.9\linewidth]{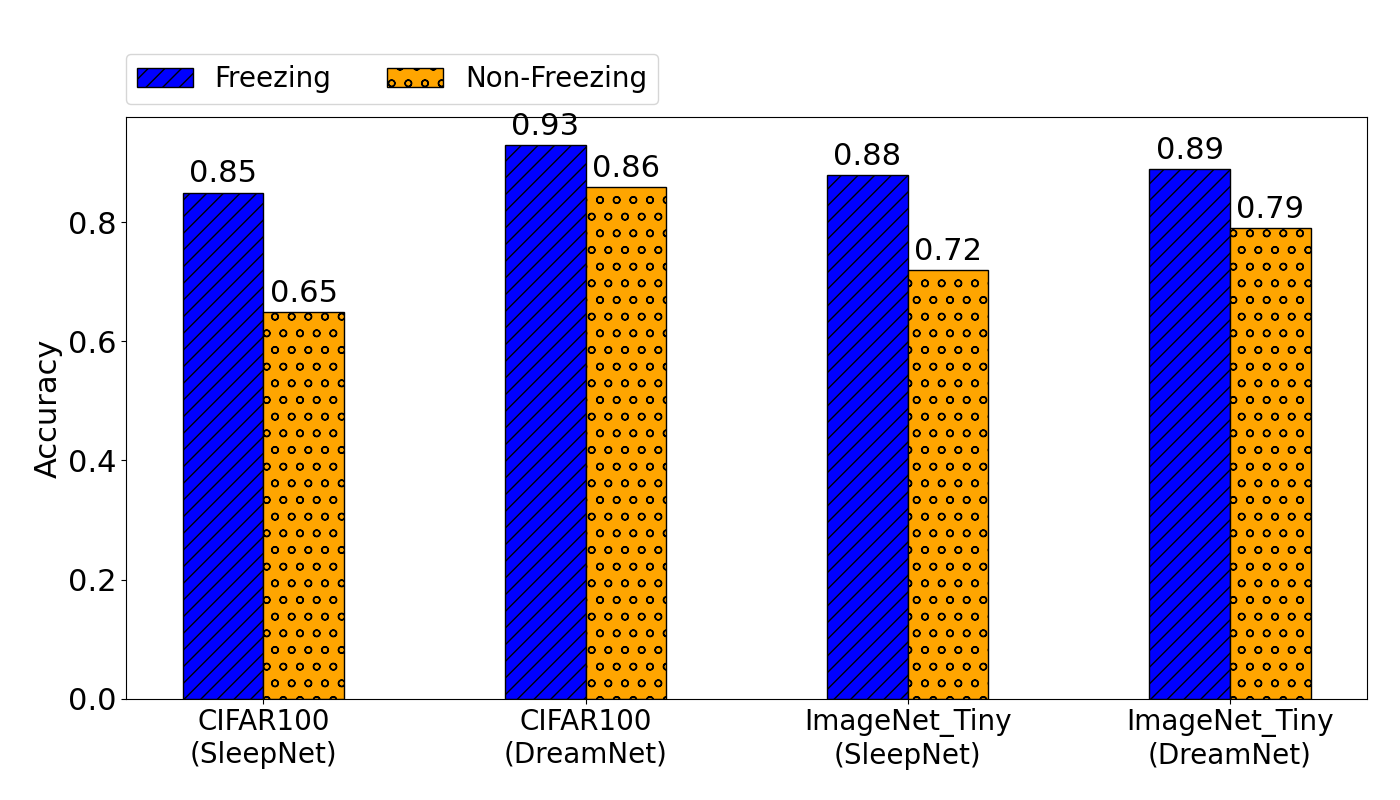}
\caption{This plot delineates the impact of freezing (marked in blue) and unfreezing (indicated in orange) the parameters of unsupervised models on CIFAR100 and ImageNet Tiny.}
\label{fig: ablation_frozen}
\end{figure}

Equally importantly, we examined the effect of freezing versus unfreezing the parameters of pre-trained encoders and autoencoders. This evaluation was carried out for image classification on CIFAR100, employing ResNet18 coupled with Google's ViT Base. Fig \ref{fig: ablation_frozen} demonstrates that consistently, the ``Frozen'' configuration (depicted in blue) outperforms the ``Non-Frozen'' one (shown in orange). Our proposed models with frozen encoders generally achieved higher accuracy for both datasets.

We attribute this consistently superior performance of frozen encoders to two major reasons. Firstly, unfreezing the parameters during training may often tend to overfit the model. The overfitting can be traced back to finding an optimal learning rate for such a setup. More specifically, the supervised component of the model, which is not pre-trained, requires a larger learning rate to capture complex patterns effectively. In contrast, the pre-trained unsupervised part necessitates a lower learning rate to avoid drastic changes that can degrade the valuable pre-trained patterns. Balancing this diverse learning rate needs while unfreezing parameters is non-trivial and often leads to overfitting. Secondly, since the unsupervised models contribute additional features to the supervised models, any parameter alteration could modify these supplementary features to fit the specific dataset being processed. Although this might seem beneficial, it could inadvertently filter out some of the generalized, useful latent information that the unsupervised encoder initially captured, thereby limiting the model's overall ability to generalize across diverse datasets.

\subsection{Complexity and Qualitative Results}
Experiments were conducted on a RHEL 7.9 system with an Intel Xeon Gold 6238R CPU, NVIDIA Quadro RTX 5000 GPU, and 88GB RAM. Table \ref{tab: efficiency} compares training time per epoch on CIFAR100 and ImageNet-tiny. ResNet18 was the fastest, requiring 0.45 and 6.77 hours per epoch, respectively. SleepNet was also efficient, outperforming ViT-B with 0.80 vs.\ 1.04 hours on CIFAR100 and 8.81 vs.\ 10.60 hours on ImageNet-tiny, indicating strong computational efficiency.

From Table \ref{tab: cv model_performance}, ResNet18 had the lowest complexity at 1.8B FLOPs, while DreamNet-4 reached 50.5B FLOPs and achieved higher accuracy, with 92.3\% on CIFAR100 and 89.9\% on ImageNet-tiny. This reflects the trade-off between complexity and performance. Overall, SleepNet and DreamNet provide a good balance between efficiency and accuracy. While DreamNet is more computationally demanding, it delivers clear performance gains, showing that its added architectural complexity is effective for computer vision tasks.
\begin{table}[t]
    \centering
    \caption{Efficiency is evaluated by the hour per epoch on CIFAR100 and ImageNet Tiny. The best-performed model is highlighted in bold. EffNetB7 refers to EfficientNet-B7.}
    \begin{tabular}{@{}p{2.5cm}@{}p{1.2cm}@{}p{1.2cm}@{}p{1.2cm}@{}p{1.2cm}@{}p{1.2cm}@{}p{1.2cm}@{}p{1.2cm}@{}}
    \toprule
        \textbf{Datasets} & \rotatebox{45}{ResNet18} & \rotatebox{45}{ResNet50} & \rotatebox{45}{EffNetB7} & \rotatebox{45}{\(\text{ViT}_{\text{base}}\)} & \rotatebox{45}{\(\text{MAE}_{\text{base}}\)} & \rotatebox{45}{SleepNet} & \rotatebox{45}{DreamNet} \\ \midrule

        CIFAR100 & \textbf{0.45} & 0.65 & 0.55 & 1.04 & 0.73 & 0.80 & 1.80 \\ \midrule
        \shortstack{ImageNet\\Tiny} & \textbf{6.77} & 7.43 & 7.11 & 10.60 & 8.60 & 8.81 & 18.90 \\ 
        \bottomrule
    \end{tabular}
\label{tab: efficiency}
\end{table}
\section{Conclusion}\label{conclusion}
In this paper, we introduced two innovative deep learning architectures, SleepNet and DreamNet. These models use novel enrichment and reconstruction mechanisms to consolidate and refine visual features, leading to improved performance in computer vision tasks. Our experiments show that these models outperform state-of-the-art results, highlighting their effectiveness and potential for general applicability.

\begin{credits}

\subsubsection{\discintname}
The authors declared that they have no conflicts of interest in this work.

\end{credits}
%
%
%
\bibliographystyle{splncs04}
\bibliography{ref}

\end{document}